\documentclass[10pt,twocolumn,letterpaper]{article}

\usepackage[pagenumbers]{cvpr} 

\usepackage{graphicx}
\usepackage{amsmath}
\usepackage{amssymb}
\usepackage{booktabs}
\usepackage{makecell}
\usepackage{multirow}
\usepackage{colortbl}
\usepackage{tabu}
\usepackage[accsupp]{axessibility}  
\usepackage{pifont}
\newcommand{\cmark}{\ding{51}\xspace}%
\newcommand{\xmarkg}{\textcolor{lightgray}{\ding{55}}\xspace}%
\usepackage[misc]{ifsym}
\usepackage{flushend}

\usepackage[pagebackref,breaklinks,colorlinks]{hyperref}

\usepackage[capitalize]{cleveref}
\crefname{section}{Sec.}{Secs.}
\Crefname{section}{Section}{Sections}
\Crefname{table}{Table}{Tables}
\crefname{table}{Tab.}{Tabs.}

\newcommand{\Rmnum}[1]{\expandafter\@slowromancap\romannumeral #1@}

\usepackage{color}
\usepackage{xcolor}

\newcommand{\ntacc}{N-acc.\xspace}
\newcommand{\tacc}{T-acc.\xspace}

\let\oldsubsection\subsection
\renewcommand{\subsection}[1]{\oldsubsection{#1} }

\def\cvprPaperID{1139} 
\def\confName{CVPR}
\def\confYear{2023}

\begin{document}

\title{GRES: Generalized Referring Expression Segmentation}
\author{
Chang Liu\footnotemark[2]
\qquad
Henghui Ding\footnotemark[2]~~$^{\textrm{\Letter}}$
\qquad
Xudong Jiang\\
Nanyang Technological University, Singapore\\
\href{https://henghuiding.github.io/GRES}{https://henghuiding.github.io/GRES}
}
\vspace{-5mm}

\twocolumn[{%
\renewcommand\twocolumn[1][]{#1}%
\maketitle 
\vspace{-5mm}
\begin{center} 
\centering 
\includegraphics[width=1\textwidth]{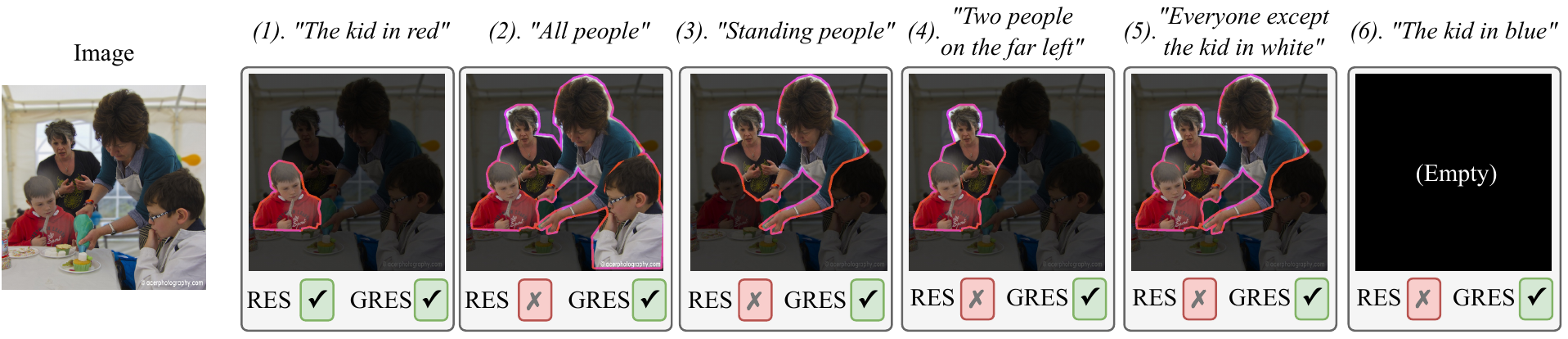} 
\vspace{-6mm}
\captionof{figure}{Classic Referring Expression Segmentation (RES) only supports expressions that indicate a single target object, \eg, (1). Compared with classic RES, the proposed \textbf{Generalized Referring Expression Segmentation (GRES)} supports expressions indicating an
\textbf{\textit{arbitrary number}} of target objects, for example, multi-target expressions like ({2})-({5}), and no-target expressions like ({6}).}
\label{fig:fig1} 
\end{center}%
}]

\renewcommand{\thefootnote}{\fnsymbol{footnote}}
\footnotetext[2]{Equal contribution.}
\footnotetext[0]{${\textrm{\Letter}}$ Corresponding author (henghui.ding@gmail.com).}

\begin{abstract}
  Referring Expression Segmentation (RES) aims to generate a segmentation mask for the object described by a given language expression. Existing classic RES datasets and methods commonly support single-target expressions only, \ie, one expression refers to one target object. Multi-target and no-target expressions are not considered. This limits the usage of RES in practice. In this paper, we introduce a new benchmark called Generalized Referring Expression Segmentation (GRES), which extends the classic RES to allow expressions to refer to an arbitrary number of target objects. Towards this, we construct the first large-scale GRES dataset called gRefCOCO that contains multi-target, no-target, and single-target expressions. GRES and gRefCOCO are designed to be well-compatible with RES, facilitating extensive experiments to study the performance gap of the existing RES methods on the GRES task. In the experimental study, we find that one of the big challenges of GRES is complex relationship modeling. Based on this, we propose a region-based GRES baseline ReLA that adaptively divides the image into regions with sub-instance clues, and explicitly models the region-region and region-language dependencies. The proposed approach ReLA achieves new state-of-the-art performance on the both newly proposed GRES and classic RES tasks. The proposed gRefCOCO dataset and method are available at \href{https://henghuiding.github.io/GRES}{https://henghuiding.github.io/GRES}.
  \vspace{-4mm}
\end{abstract}

\section{Introduction}
\label{sec:intro}

Referring Expression segmentation (RES) is one of the most important tasks of multi-modal information processing. Given an image and a natural language expression that describes an object in the image, RES aims to find this target object and generate a segmentation mask for it. It has great potential in many applications, such as video production, human-machine interaction, and robotics. Currently, most of the existing methods follow the RES rules defined in the popular datasets ReferIt \cite{kazemzadeh-etal-2014-referitgame} and RefCOCO \cite{yu2016modeling,mao2016generation} and have achieved great progress in recent years. 

\textbf{Limitations of classic RES.} However, most classic RES methods have some strong pre-defined constraints to the task. First, the classic RES does not consider no-target expressions that do not match any object in the image. This means that the behavior of the existing RES methods is undefined if the target does not exist in the input image. When it comes to practical applications under such constraint, the input expression has to match an object in the image, otherwise problems inevitably occur. Second, most existing datasets, \eg, the most popular RefCOCO \cite{yu2016modeling,mao2016generation}, do not contain multi-target expressions that point to multiple instances. This means that multiple 
inputs are needed to search objects one by one. \Eg, in \cref{fig:fig1}, four distinct expressions with four times of model calls are required to segment \textit{``All people''}. Our experiments show that classic RES methods trained on existing datasets cannot be well-generalized to these scenarios.

\begin{table}[t]
  \renewcommand\arraystretch{1.2}
  \centering
  \footnotesize
  \caption{Comparison among different referring expression data-sets, including ReferIt\cite{kazemzadeh-etal-2014-referitgame}, RefCOCO(g)\cite{yu2016modeling,mao2016generation}, PhraseCut\cite{wu2020phrasecut}, and our proposed \textbf{gRefCOCO}. Multi-target: expression that specifies multiple objects in the image. No-target: expression that does not touch on any object in the image.}\vspace{-3mm}
  \setlength{\tabcolsep}{1.36mm}{\begin{tabular}{lcccc}
    \specialrule{.1em}{.05em}{.05em} 
          & ReferIt & RefCOCO(g)& PhraseCut&  \textbf{gRefCOCO}\\
          \cline{2-5}
          Image Source  & CLEF\cite{grubinger2006iapr} & COCO\cite{lin2014microsoft} & VG\cite{krishna2017visual}  & COCO\cite{lin2014microsoft} \\
          Multi-target & {\xmarkg} & {\xmarkg} & {(fallback)} & {\cmark} \\
          No-target & {\xmarkg} & {\xmarkg} & {\xmarkg}  & {\cmark} \\
          \makecell[c]{Expression type} & free  & free  & templated & free \\
    \specialrule{.1em}{.05em}{.05em} 
    \end{tabular}}%
  \label{tab:dataset_compare}%
\vspace{-5mm}
\end{table}%
\textbf{New benchmark and dataset.} In this paper, we propose a new benchmark, called Generalized Referring Expression Segmentation (GRES), which allows expressions indicating any number of target objects. GRES takes an image and a referring expression as input, the same as classic RES. Different from classic RES, as shown in \cref{fig:fig1}, GRES further supports multi-target expression that specifies multiple target objects in a single expression, \eg, \textit{``Everyone except the kid in white''}, and no-target expression that does not touch on any object in the image, \eg, \textit{``the kid in blue''}. This provides much more flexibility for input expression, making referring expression segmentation more useful and robust in practice. However, existing referring expression datasets~\cite{kazemzadeh-etal-2014-referitgame,yu2016modeling,mao2016generation} do not contain multi-target expression nor no-target samples, but only have single-target expression samples, as shown in \cref{tab:dataset_compare}. To facilitate research efforts on realistic referring segmentation, we build a new dataset for GRES, called gRefCOCO. It complements RefCOCO with two kinds of samples: multi-target samples, in which the expression points to two or more target instances in the image, and no-target samples, in which the expression does not match any object in the image.

\textbf{A baseline method.} Moreover, we design a baseline method based on the objectives of the GRES task. It is known that modeling relationships, \eg, region-region interactions, plays a crucial role in RES \cite{yu2018mattnet}. However, classic RES methods
only have one target to detect so that many methods can achieve good performance without explicit region-region interaction modeling. But in GRES, as multi-target expressions involve multiple objects in one expression, 
it is more challenging and essential to model the long-range region-region dependencies. 
From this point, we propose a region-based method for GRES that explicitly model the interaction among regions with sub-instance clues. 
We design a network that splits the image into regions and makes them explicitly interact with each other.
Moreover, unlike previous works where regions come from a simple hard-split of the input image, our network soft-collates features for each region, achieving more flexibility. We do extensive experiments on our proposed methods against other RES methods, showing that the explicit modeling of interaction and flexible region features greatly contributes to the performance of GRES. 

In summary, our contributions are listed as follows:
\vspace{-1mm}
\begin{enumerate}
\setlength\itemsep{0em}
  \item We propose a benchmark of Generalized Referring Expression Segmentation (GRES), making RES more flexible and practical in real-world scenarios.
  \item We propose a large-scale GRES dataset gRefCOCO. To the best of our knowledge, this is the first referring expression dataset that supports expressions indicating an arbitrary number of target objects.
  \item We propose a solid baseline method ReLA for GRES to model complex \textbf{ReLA}tionships among objects, which achieves the new state-of-the-art performance on {{both classic RES and newly proposed GRES}} tasks.
  \item We do extensive experiments and comparisons of the proposed baseline method and other existing RES methods on the GRES, and analyze the possible causes of the performance gap and new challenges in GRES.
\end{enumerate}

\begin{figure*}[t]
  \centering
  \vspace{0.35em}
  \hfill
  \begin{subfigure}[t]{0.5\linewidth}
      \centering
      \includegraphics[width=\textwidth]{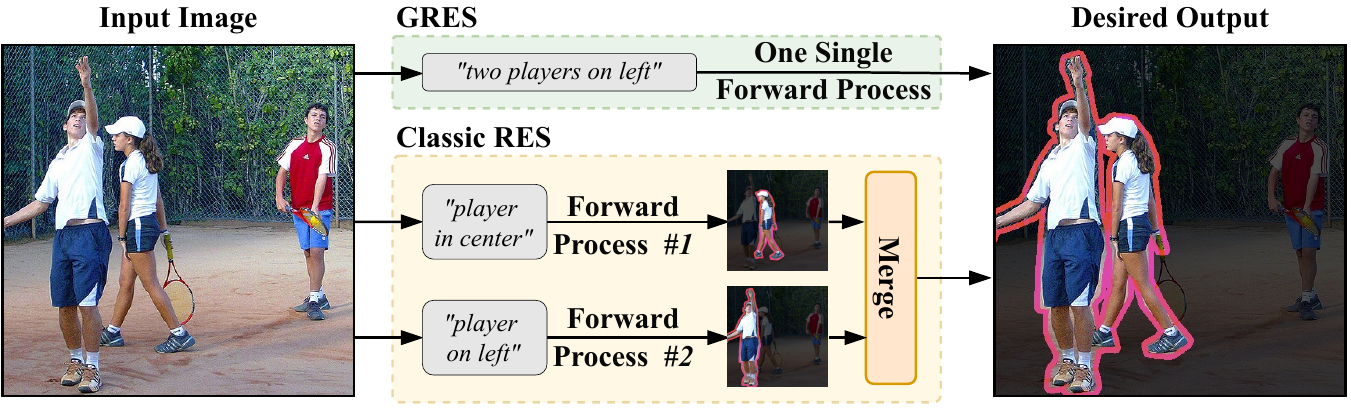}
      \caption{Multi-target: Selecting multiple objects in one single forward process.}
      \label{fig:app_mt}
  \end{subfigure}
  \hfill
  \begin{subfigure}[t]{0.47\linewidth}
      \centering
      \includegraphics[width=\textwidth]{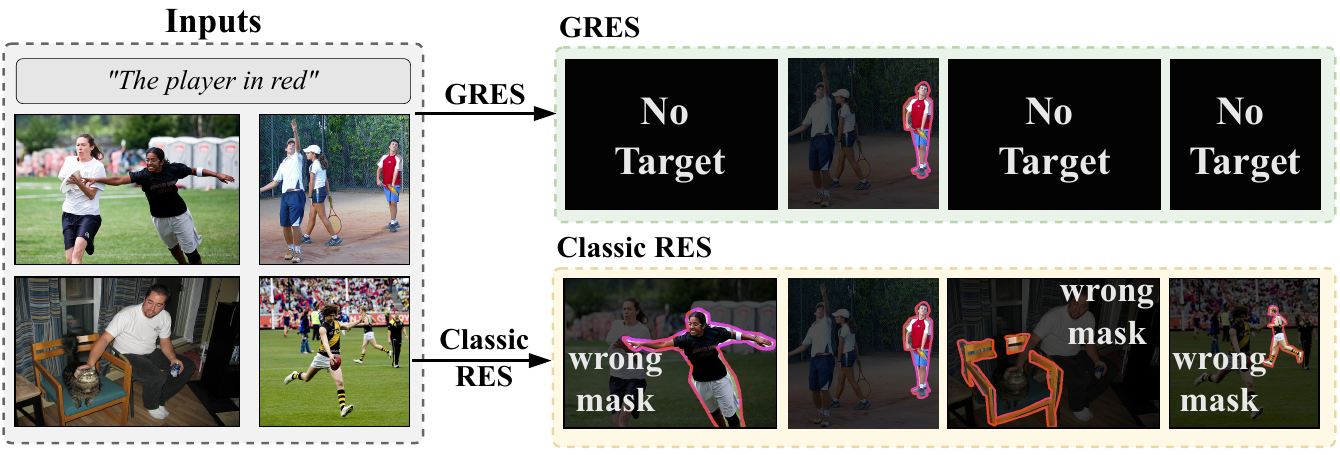}
      \caption{No-target: Retrieving images that contain the object.}
      \label{fig:app_nt}
  \end{subfigure}
  \hfill
 \vspace{-0.08in}
  \caption{More applications of GRES brought by supporting multi-target and no-target expressions compared to classic RES.}
  \vspace{-0.15in}
  \label{fig:gres_app}
\end{figure*}

\section{Related Works} \label{sec:related_works}
\textbf{Related referring tasks and datasets.} Being defined by Hu \etal \cite{hu2016segmentation}, Referring Expression \textit{Segmentation} (RES) comes from a similar task, Referring Expression \textit{Comprehension} (REC) \cite{hu2016natural,wang2019neighbourhood,liu2019learning,yang2019fast,zhuang2018parallel,yang2020improving,liao2020real} that outputs a bounding box for the target. The earliest dataset for RES and REC is ReferIt \cite{kazemzadeh-etal-2014-referitgame}
, in which one expression only refers to one instance. Later, Yu \etal propose RefCOCO \cite{yu2016modeling} for RES and REC. However, like ReferIt, it only contains single-target expressions. Another popular dataset RefCOCOg \cite{mao2016generation} also inherits this. Although the original definition of RES \cite{hu2016segmentation} does not limit the number of target instances, ``\textbf{\textit{one expression, one instance}}'' has become a ``de-facto'' rule for RES task. 

Recently, some new datasets are proposed, but most of them are neither focused on nor suitable for GRES. \Eg, although PhraseCut \cite{wu2020phrasecut} has multi-target expressions, it only considers them as ``fallback'', \ie, multi-target expressions are only used when an object cannot be uniquely referred to. In contrast, our expression intentionally finds multiple targets. Besides, expressions in PhraseCut are written using templates rather than free natural language expressions, limiting the diversity of language usage. Image caption datasets~\cite{plummer2015flickr30k,krishna2017visual} are close to RES, but they cannot ensure unambiguity of expression$\rightarrow$object(s).
Thus, they are not suitable for referring-related tasks. There are some referring datasets using other data modalities or learning schemes, \eg, Scanrefer \cite{chen2020scanrefer} focuses on 3D objects and Clevrtex \cite{karazija2021clevrtex} focuses on unsupervised learning. Moreover, none of the above datasets has no-target expression.

\textbf{Referring segmentation methods.} Referring segmentation methods can be roughly divided into two categories: one-stage (or top-down) methods~\cite{margffoy2018dynamic,zhang2022coupalign,li2018referring,chen2019see,ye2019cross,hu2020bi,huang2020referring,hui2020linguistic,luo2020cascade} and two-stage (or bottom-up) methods~\cite{yu2018mattnet,liu2022instance}. One-stage methods usually have an FCN-like \cite{long2015fully} end-to-end network, and the prediction is achieved by per-pixel classification on fused multi-modal feature. 
Two-stage methods first find a set of instance proposals using an out-of-box instance segmentation network and then select the target instance from them. The majority of RES methods are one-stage, while two-stage methods are more prevalent in REC~\cite{luo2017comprehension,hu2017modeling,hu2016natural,liu2017referring,yu2017joint,zhang2017discriminative}. Most recently, some transformer-based methods~\cite{VLTPAMI,MOSE,wang2022cris,M3Att,li2023transformer} are proposed and bring large performance gain compared to the CNN-based network. 
Zero-shot segmentation methods~\cite{FZShot3D,PAD,zhanghui2021} use class names as textual information and focus on identifying novel categories, in contrast to RES that employs natural expressions to identify the user's intended target.

\section{Task Setting and Dataset}

\subsection{GRES Settings}
\textbf{Revisit of RES.} Classic Referring Expression Segmentation (RES) takes an image and an expression as inputs. The desired output is a segmentation mask of the target region that is referred by the input expression. As discussed in \cref{sec:related_works}, the current RES does not consider no-target expressions, and
all samples in current datasets only have single-target expressions. Thus, existing models are likely to output an instance incorrectly if the input expression refers to nothing or multiple targets in the input image.

\textbf{Generalized RES.} To address these limitations in classic RES, we propose a benchmark called Generalized Referring Expression Segmentation (GRES) that allows expressions indicating arbitrary number of target objects. A GRES data sample contains four items: an image $I$, a language expression $T$, a ground-truth segmentation mask $M_\mathit{GT}$ that covers pixels of all targets referred by $T$, and a binary no-target label $E_\mathit{GT}$ that indicates whether $T$ is a no-target expression. 
The number of instances in $T$ is not limited. GRES models take $I$ and $T$ as inputs and predict a mask $M$.
For no-target expressions, $M$ should be all negative.

The applications of multi-target and no-target expressions are not only finding multiple targets and rejecting inappropriate expressions matching nothing, but also bringing referring segmentation into more realistic scenarios with advanced usages. For example, with the support of multi-target expressions, we can use expressions like ``\textit{all people}'' and ``\textit{two players on left}'' as input to select multiple objects in a single forward process (see \cref{fig:app_mt}), or use expressions like ``\textit{foreground}'' and ``\textit{kids}'' to achieve user-defined open vocabulary segmentation. With the support of no-target expressions, users can apply the same expression on a set of images to identify which images contain the object(s) in the language expression, as in \cref{fig:app_nt}. This is useful if users want to find and matte something in a group of images, similar to image retrieval but more specific and flexible. What's more, allowing multi-target and no-target expressions enhances the model's reliability and robustness to realistic scenarios where any type of expression can occur unexpectedly, for example, users may accidentally or intentionally mistype a sentence. 

\textbf{Evaluation.} To encourage the diversity of GRES methods, we do not force GRES methods to differentiate different instances in the expression though our dataset gRefCOCO provides, enabling popular one-stage methods to participate in GRES. Besides the regular RES performance metric cumulative IoU (cIoU) and Precision@X, we further propose a new metric called generalized IoU~(gIoU), which extends the mean IoU to all samples including no-target ones. Moreover, No-target performance is also separately evaluated by computing No-target-accuracy (\ntacc)~and Target-accuracy~(\tacc). Details are given in \cref{sec:metrics}.

\subsection{gRefCOCO: A Large-scale GRES Dataset}

\begin{figure*}[t]
  \begin{center}
     \includegraphics[width=0.996\linewidth]{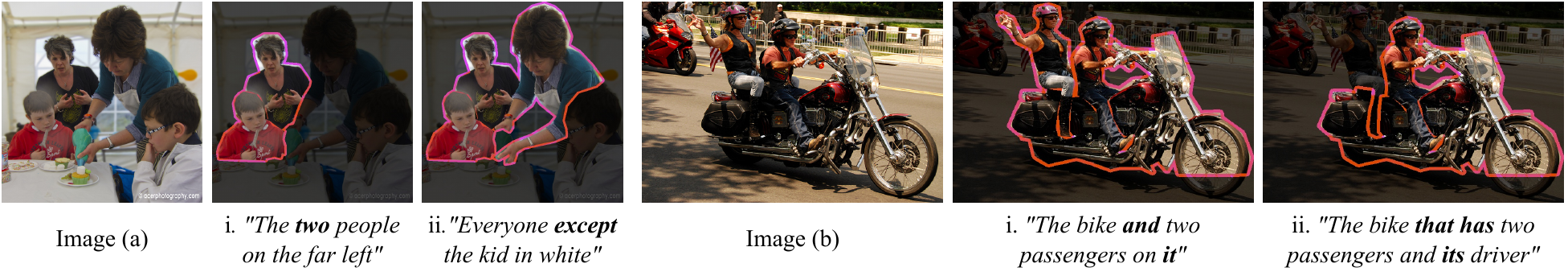}
  \end{center}
 \vspace{-0.26in}
  \caption{Examples of the proposed gRefCOCO dataset.}
  \vspace{-0.15in}
  \label{fig:dataset_example}
\end{figure*}

To perform the GRES task, we construct the gRefCOCO dataset.
It 
contains 278,232 expressions, including 80,022 multi-target and 32,202 no-target expressions, referring to 60,287 distinct instances in 19,994 images. 
Masks and bounding boxes for all target instances are given.
Part of single-target expressions are inherited from RefCOCO. 
We developed an online annotation tool to find images, select instances, write expressions, and verify the results.

The basic annotation procedure follows ReferIt \cite{kazemzadeh-etal-2014-referitgame} to ensure the annotation quality. The data split is also kept the same as the UNC partition of RefCOCO \cite{yu2016modeling}. 
We compare the proposed gRefCOCO with RefCOCO and list some unique and significant features of our dataset as follows.

{\textbf{Multi-target samples.}} 
In practice, users usually cluster multiple targets of an image by describing their logical relationships or similarities. {From this point, we let annotators select target instances rather than randomly assembling them. Then annotators write an unambiguous referring expression for the selected instances.} 
There are four major features and challenges brought by multi-target samples:  

\vspace{2pt}\noindent\textbf{1) Usage of counting expressions}, \eg, ``\textit{The \textbf{two} people on the far left}'' in \cref{fig:dataset_example}(a). As the original RefCOCO already has ordinal word numbers like ``\textit{the \textit{second} person from left}'', the model must be able to differentiate cardinal numbers from ordinal numbers. Explicit or implicit object-counting ability is desired to address such expressions.

\vspace{2pt}\noindent\textbf{2) Compound sentence structures without geometrical relation}, like compound sentences \textit{``A \textbf{and} B''}, \textit{``A \textbf{except} B''}, and \textit{``A \textbf{with} B \textbf{or} C''}, as shown in \cref{fig:dataset_example}. This raises higher requirements for models to understand the long-range dependencies of both the image and the sentence.

\vspace{2pt}\noindent\textbf{3) Domain of attributes.} When there are multiple targets in an expression, different targets may share attributes or have different attributes, \eg, \textit{``the right lady in blue and kid in white''}. Some attributes may be shared, \eg, \textit{``right''}, and others may not, \eg, \textit{``blue''} and \textit{``white''}.
This requires the model to have a deeper understanding of all the attributes and map the relationship of these attributes to their corresponding objects. 

 \vspace{2pt}\noindent\textbf{4) More complex relationships}. Since a multi-target expression involves more than one target, relationship descriptions appear more frequently and are more complicated than in sing-target ones. 
\cref{fig:dataset_example}(b) gives an example. 
Two similar expressions are applied on the same image. Both expressions have the conjunction word \textit{``and''}, and \textit{``two passengers''} as an attribute to the target \textit{``bike''}. 
But the two expressions refer to two different sets of targets as shown in \cref{fig:dataset_example}(b).
Thus in GRES, relationships are not only used to describe the target but also indicate the number of targets. This requires the model to have a deep understanding of all instances and their interactions in the image and expression.

 \vspace{2pt} \textbf{No-target samples.} {During the annotation, we found that if we do not set any constraints for no-target expressions, annotators tend to write a lot of simple or general expressions that are quite different from other expressions with valid targets. \Eg, annotators may write duplicated ``\textit{dog}'' for all images without dogs. 
To avoid these undesirable and purposeless samples in the dataset,
we set two rules for no-target expressions:}

\noindent\textbf{1) The expression cannot be totally irrelevant to the image}. For example, given the image in \cref{fig:dataset_example}(a), ``\textit{The kid in blue}'' is acceptable as there do exist kids in the image, but none of them is in blue. But expressions like ``\textit{dog}'', ``\textit{car}'', \textit{``river''} \etc are unacceptable as they are totally irrelevant to anything in this image. 

\vspace{2pt}\noindent\textbf{2) The annotators could choose a deceptive expression} drawn from other images in RefCOCO's same data split, if an expression required by in 1) is hard to come up with.

These rules greatly improve the diversity of no-target expressions and keep our dataset at a reasonable difficulty. More examples are shown in the Supplementary Materials.

\begin{figure*}[t]
  \begin{center}
     \includegraphics[width=0.996\linewidth]{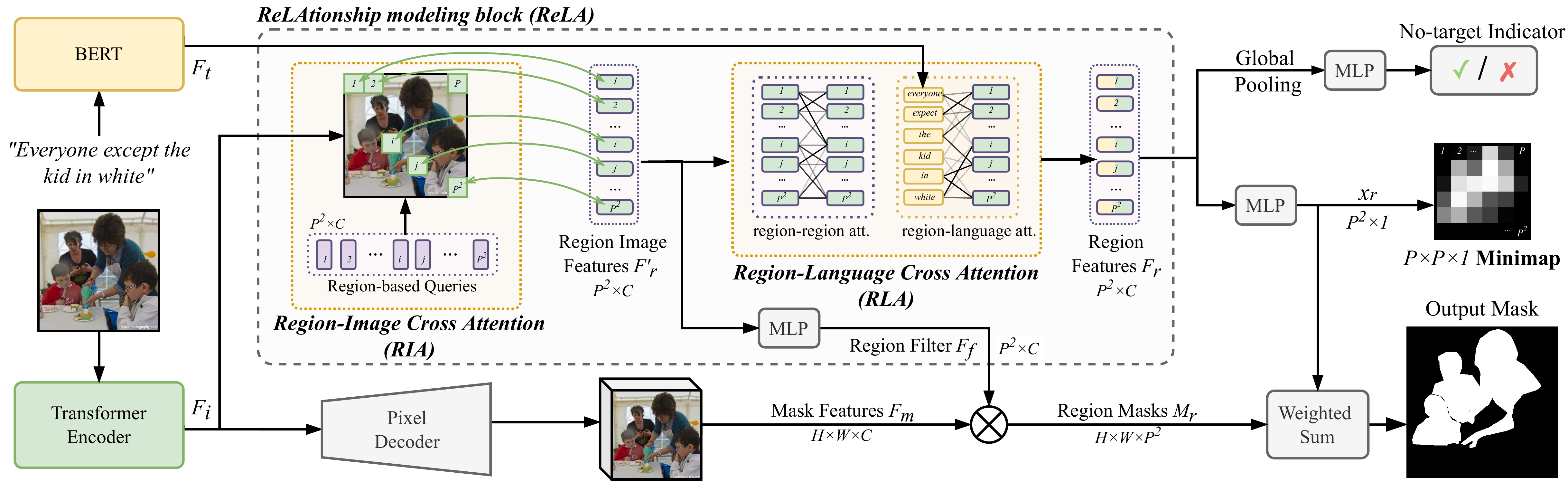}
  \end{center}
 \vspace{-0.17in}
  \caption{Architecture overview of the GRES baseline model \textbf{ReLA}. Firstly, the given image and expression are encoded into vision feature $F_i$ and language feature $F_t$, respectively. $F_i$ is fed into a pixel decoder to produce mask features $F_m$. \textbf{ReLA}tionship modeling block takes both $F_i$ and $F_t$ as inputs and output 1) region filter $F_f$ that produces region masks $M_r$, 2) region probability map $x_r$, and 3) no-target judgement score $E$. Output mask is obtained by weighted fusion of region masks $M_r$.}
  \vspace{-0.1in}
  \label{fig:net_arch}
\end{figure*}
\section{The Proposed Method for GRES}

As discussed earlier, the relationship and attribute descriptions are more complex in multi-target expressions. Compared with classic RES, it is more challenging and important for GRES to model the complex interaction among regions in the image, and capture fine-grained attributes for all objects. We propose to explicitly interact different parts of image and different words in expression to analyze their dependencies.

\subsection{Architecture Overview}

The overview of our framework is shown in \cref{fig:net_arch}. The input image is processed by a transformer encoder based on Swin \cite{liu2021swin} to extract vision features $F_i\in \mathbb{R}^{H\times W\times C}$, in which $H, W$ are the spatial size and $C$ is the channel dimensions. The input language expression is processed by BERT \cite{devlin2018bert}, producing the language feature $F_t\in \mathbb{R}^{N_t\times C}$, where $N_t$ is the number of words in the expression. 
Next, $F_i$ is sent to a pixel decoder to obtain the mask feature $F_m$ for mask prediction. Meantime, $F_i$ and $F_t$ are sent to our proposed \textbf{ReLA}tionship modeling block (see \cref{sec:ReLA} for details), which divides the feature maps into $P\times P\!=\!P^2$ regions, and models the interaction among them. These ``regions'' correspond to the image's $P\times P$ patches like ViT\cite{dosovitskiy2020image}. However, the shape and sizes of their spatial areas are not predefined but found by ReLA dynamically, which is different from previous works using hard-split~\cite{dosovitskiy2020image,xie2021segformer,strudel2021segmenter,kim2022restr}. 
ReLA generates two sets of features: region feature $F_r\!=\!\{f_r^n\}_{n=1}^{P^2}$ and region filter $F_f\!=\!\{f_f^n\}_{n=1}^{P^2}$. For the $n$-th region, its region feature $f_r^n$ is used to find a scalar $x_r^n$ that indicates its probability of containing targets, and its region filter $f_f^n$ is multiplied with the mask feature $F_m$, generating its regional segmentation mask $M_r^n\in \mathbb{R}^{H\times W}$ that indicates the area of this region. We get the predicted mask by weighted aggregating these masks: 
\vspace{-2mm}\begin{equation}\vspace{-3mm}
  M = \sum_n(x_r^n M_r^n).
\end{equation}

\textbf{Outputs and Loss.} The predicted mask $M$ is supervised by the ground-truth target mask $M_\mathit{GT}$. The $P\times P$ probability map $x_r$ is supervised by a ``minimap'' downsampled from $M_\mathit{GT}$, so that we can link each region with its corresponding patch in the image. Meantime, we take the global average of all region features $F_r$ to predict a no-target label $E$. In inference, if $E$ is predicted to be positive, the output mask $M$ will be set to empty. $M$, $x_r$ and $E$ are guided by the cross-entropy loss.

\subsection{ReLAtionship Modeling}\label{sec:ReLA}

The proposed \textbf{ReLA}tionship modeling has two main modules, Region-Image Cross Attention (RIA) and Region-Language Cross Attention (RLA). The RIA flexibly collects region image features. The RLA captures the region-region and region-language dependency relationships.

\textbf{Region-Image~Cross~Attention~(RIA).} 
RIA takes the vision feature $F_i$ and $P^2$ learnable Region-based Queries $Q_{r}$ as input. Supervised by the minimap shown in \cref{fig:net_arch}, each query corresponds to a spatial region in the image and is responsible for feature decoding of the region. 
The architecture is shown in \cref{sub@fig:ria}. First, the attention between image feature $F_i$ and $P^2$ query embeddings $Q_{r}\in \mathbb{R}^{P^2\times C}$ is performed to generate $P^2$ attention maps:
\vspace{-1mm}\begin{equation}\vspace{-1mm}
  A_{ri} = \text{softmax}(Q_{r}\sigma (F_{i}W_{ik})^T),
\label{eq:a-ri}
\end{equation}
where $W_{ik}$ is $C\!\times\!C$ learnable parameters and $\sigma$ is GeLU~\cite{hendrycks2016gaussian}. The resulting $A_{ri}\in\mathbb{R}^{P^2\times HW}$ gives each query a $H\!\times\!W$ attention map indicating its corresponding spatial areas in the image. Next, we get the region features from their corresponding areas using these attention maps: $F'_r=A_{ri} \sigma (F_{i}W_{iv})^T$, where $W_{iv}$ is $C\!\times\!C$ learnable parameters.
In such a way, the feature of each region is dynamically collected from their relevant positions. Compared to hard-splitting the image into patches, this method gives more flexibility. An instance may be represented by multiple regions in the minimap (see \cref{fig:net_arch}), making regions represent more fine-grained attributes at the sub-instance level, \eg, the head and upper body of a person. Such sub-instance representations are desired for addressing the complex relationship and attribute descriptions in GRES. A region filter $F_f$ containing region clues is obtained based on $F'_r$ for mask prediction. $F'_r$ is further fed into RLA for region-region and region-word interaction modeling.

\begin{figure}[t]
  \centering
  \vspace{0.35em}
  \hfill
  \begin{subfigure}[b]{0.493\linewidth}
      \centering
      \includegraphics[width=\textwidth]{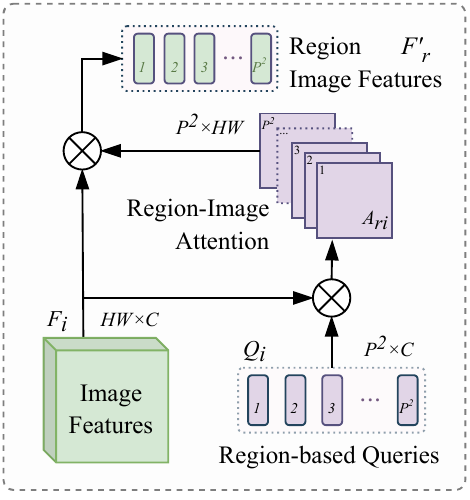}
      \caption{RIA}
      \label{fig:ria}
  \end{subfigure}
  \hfill
  \begin{subfigure}[b]{0.493\linewidth}
      \centering
      \includegraphics[width=\textwidth]{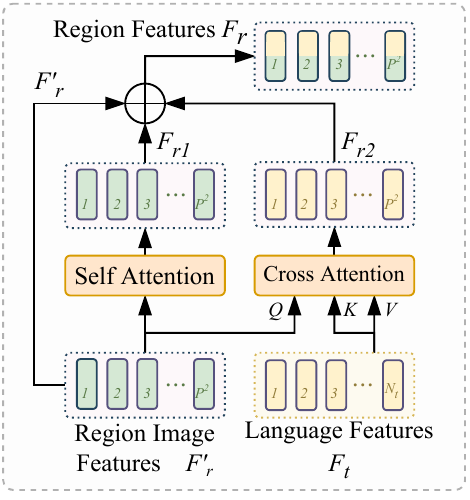}
      \caption{RLA}
      \label{fig:rla}
  \end{subfigure}
  \hfill
 \vspace{-0.3in}
  \caption{Architectures of Region-Image Cross Attention (RIA) and Region-Language Cross Attention (RLA).}
  \vspace{-0.16in}
  \label{fig:ria_rla}
\end{figure}

\textbf{Region-Language Cross Attention (RLA).} Region image features $F'_r$ come from collating image features that do not contain relationship between regions and language information. We propose RLA module to model the region-region and region-language interactions. As in \cref{sub@fig:rla}, RLA consists of a self-attention for region image features $F'_r$ and a multi-modal cross attention. The self-attention models the region-region dependency relationships. It computes the attention matrix by interacting one region feature with all other regions and outputs the relationship-aware region feature $F_{r1}$. Meanwhile, the cross attention takes language feature $F_t$ as Value and Key input, and region image feature $F'_r$ as Query input. This firstly models the relationship between each word and each region: 
\vspace{-1mm}\begin{equation}\vspace{-1mm}
A_{l} = \text{softmax}(\sigma (F'_{r}W_{lq}) \sigma (F_{t}W_{lk})^T),
\end{equation}
where $A_{l}\in \mathbb{R}^{P^2\times N_t}$. Then it forms the language-aware region features using the derived word-region attention: $F_{r2} = A_{l} F_t$. Finally, the interaction-aware region feature $F_{r1}$, language-aware region feature $F_{r2}$, and region image features $F'_{r}$ are added together, and a MLP further fuses the three sets of features: $F_r=\text{MLP}(F'_r+F_{r1}+F_{r2})$.

\section{Experiments and Discussion}

\subsection{Evaluation Metrics} \label{sec:metrics}

Besides the widely-used RES metrics cumulative IoU (cIoU) and Precision@X (Pr@X), we further introduce No-target accuracy (\ntacc), Target accuracy (\tacc), and generalized IoU (gIoU) for GRES.

\textbf{cIoU and Pr@X}. cIoU calculates the total intersection pixels over total union pixels, and Pr@X counts the percentage of samples with IoU higher than the threshold $X$. Notably, no-target samples are excluded in Pr@X. And as multi-target samples have larger foreground areas, models are easier to get higher cIoU scores. Thus, we raise the starting threshold to 0.7 for Pr@X.

\textbf{\ntacc\ and \tacc} evaluates the model's performance on no-target identification. For a no-target sample, prediction without any foreground pixels is true positive ($\mathit{TP}$), otherwise false negative ($\mathit{FN}$). Then, \ntacc~measures the model's performance on identifying no-target samples: \ntacc~= $\frac{\mathit{TP}}{\mathit{TP}+\mathit{FN}}$. \tacc~reflects how much the generalization on no-target affects the performance on target samples, \ie how many samples that have targets are misclassified as no-target: \tacc~= $\frac{\mathit{TN}}{\mathit{TN}+\mathit{FP}}$.

\textbf{gIoU.} It is known that cIoU favors larger objects \cite{yang2021lavt,wu2020phrasecut}. As multi-target samples have larger foreground areas in GRES, we introduce generalized IoU~(gIoU) that treats all samples equally. Like mean IoU, gIoU calculates the mean value of per-image IoU over all samples. For no-target samples, 
the IoU values of true positive no-target samples are regarded as 1, while IoU values of false negative samples are treated as 0.

\begin{figure}[t]
  \begin{center}
     \includegraphics[width=0.996\linewidth]{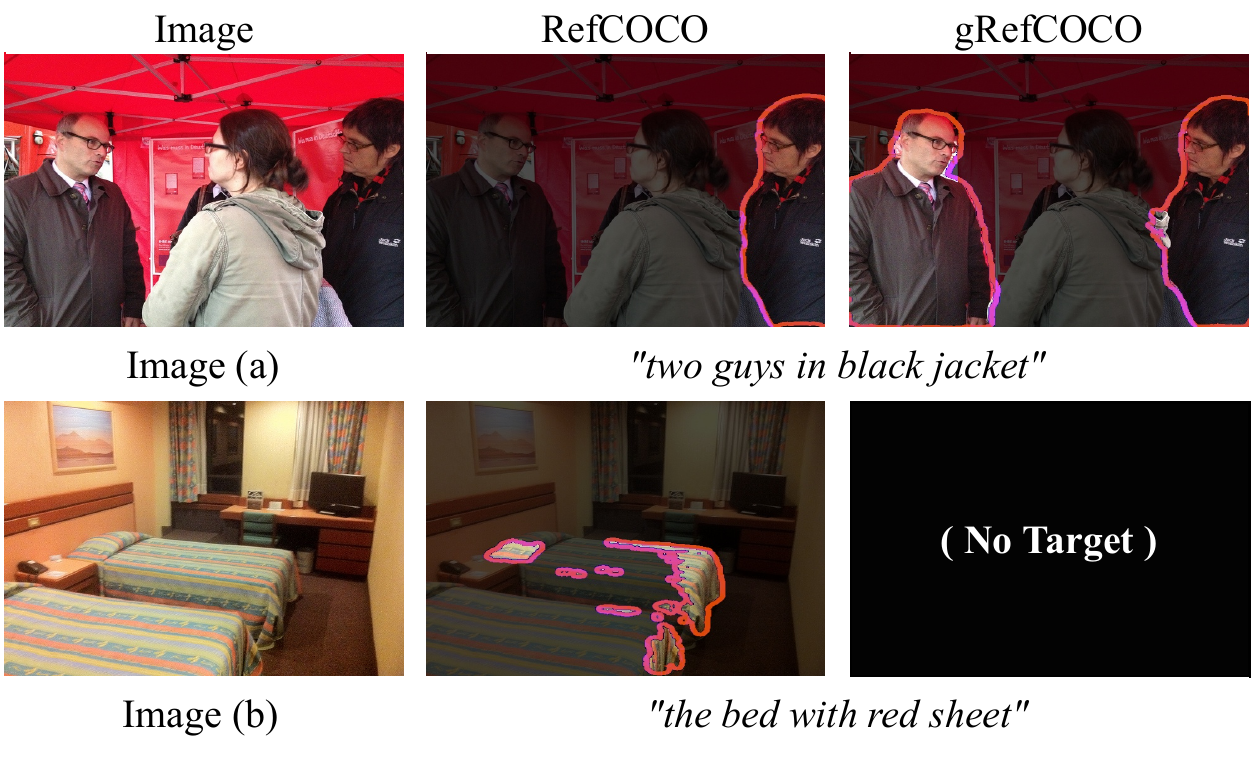}
  \end{center}
  \vspace{-0.24in}
  \caption{Example predictions of the same model being trained on RefCOCO \textit{vs.} gRefCOCO.}
  \label{fig:gres_v_res}
 \vspace{-0.05in}
\end{figure}

\subsection{Ablation Study}

\textbf{Dataset necessity.} To show the necessity and validity of gRefCOCO on the task of GRES, we compare the results of the same model trained on RefCOCO and gRefCOCO. As shown in \cref{fig:gres_v_res}, image (a) is a multi-target sample using a shared attribute (\textit{``in black jacket''}) to find \textit{``two guys''}. 
The model trained on RefCOCO only finds one, even though the expression explicitly points out that there are two target objects. Image~(b) gives a no-target expression, and the RefCOCO-trained model outputs a meaningless mask. The results demonstrate that models trained only on single-target referring expression datasets, \eg, RefCOCO, cannot be well generalized to GRES. 
In contrast, the newly built gRefCOCO can effectively enable the model to handle expressions indicating an arbitrary number of objects.

\begin{table}[t]
  \renewcommand\arraystretch{1.0}
  \centering
  \footnotesize
 \caption{Ablation study of RIA design options.}
 \vspace{-3.6mm}
 \centering
     \setlength{\tabcolsep}{1.4mm}{\begin{tabular}{r|l|ccccc}
      \specialrule{.1em}{.05em}{.05em} 
      \# & Methods  & P@0.7 & P@0.8 & P@0.9 & cIoU & gIoU\\
      \hline
      \#1 & Hard split, input  & 63.02 & 59.81 & 19.26 & 54.43 & 55.34\\
      \#2 & Hard split, decoder   & 70.34 & 65.23 & 21.47 & 60.08 & 60.93\\
      \#3 & w/o minimap   & 72.19 & 66.02 & 21.07 & 61.30 & 62.06 \\
      \hline
      \#4 & \textbf{ReLA} (ours)  & \textbf{74.20} & \textbf{68.33} & \textbf{24.68} & \textbf{62.42} & \textbf{63.60}\\
      \specialrule{.1em}{.05em}{.05em} 
   \end{tabular}}
  \label{tab:ablation_ria}
 \vspace{-0.1in}
\end{table}%

\textbf{Design options of RIA.} In \cref{tab:ablation_ria}, we investigate the performance gain brought by RIA. In model \#1, we follow previous methods \cite{dosovitskiy2020image, kim2022restr} and rigidly split the image into $P\!\times\!P$ patches before sending them into the encoder. \cref{tab:ablation_ria} shows that this method is not suitable for our ReLA framework, because it makes the global image information less pronounced due to compromised integrity. In model \#2, RIA is replaced by average pooling the image feature into $P\!\times\!P$. The gIoU gets a significant gain of $5.59\%$ from model \#1, showing the importance of global context in visual feature encoding. Then, another $2.67\%$ gIoU gain can be got by adding our proposed dynamic region feature aggregation for each query (Eq.~(\ref{eq:a-ri})), showing the effectiveness of the proposed 
adaptive region assigning.
Moreover, we study the importance of linking queries with actual image regions. In model \#3, we removed the minimap supervision so that the region-based queries $Q_r$ become plain learnable queries, resulting in a $1.54\%$ gIoU drop. This shows that explicit correspondence between queries and spatial image regions is beneficial to our network.

\begin{table}[t]
\renewcommand\arraystretch{1.0}
\centering
\footnotesize
\caption{Ablation study of RLA design options.}
 \vspace{-3.6mm}
\centering
   \setlength{\tabcolsep}{5pt}{\begin{tabular}{r|l|ccccc}
    \specialrule{.1em}{.05em}{.05em} 
    \# & Methods & P@0.7 & P@0.8 & P@0.9 & cIoU & gIoU\\
    \hline
    \#1 & Baseline  & 69.94 & 61.10 & 19.38 & 57.24 & 58.53 \\
    \#2 & + language att.  & 72.03 & 65.42 & 21.04 & 59.86 & 60.53\\
    \#3 & + region att.  & 73.52 & 67.01 & 23.43 & 61.00 & 62.38\\
    \hline
    \#4 & \textbf{ReLA} (ours)   & \textbf{74.20} & \textbf{68.33} & \textbf{24.68} & \textbf{62.42} & \textbf{63.60}\\
    \specialrule{.1em}{.05em}{.05em} 
 \end{tabular}}
\label{tab:ablation_rla}
\vspace{-0.12in}
\end{table}%

\textbf{Design options of RLA.} \cref{tab:ablation_rla} shows the importance of dependency modeling to GRES. 
In the baseline model, RLA is replaced by point-wise multiplying region features and globally averaged language features, to achieve a basic feature fusion like previous works \cite{ding2021vision, luo2020multi}. In model \#2, the language cross attention is added onto the baseline model, which brings a gIoU gain of $2\%$. This shows the validity of region-word interaction modeling. Then we further add the region self-attention to investigate the importance of the region-region relationship. The region-region relationship modeling brings a performance gain of $3.85\%$ gIoU. The region-region and region-word relationship modeling together bring a significant improvement of $5.07\%$ gIoU.

\begin{table}[t]
\renewcommand\arraystretch{1.0}
\centering
\footnotesize
\caption{Ablation study of Number of Regions}
\vspace{-3.6mm}
\centering
   \setlength{\tabcolsep}{7.76pt}{\begin{tabular}{c|ccccc}
    \specialrule{.1em}{.05em}{.05em} 
    \#~Regions & P@0.7 & P@0.8 & P@0.9 & cIoU & gIoU\\
    \hline
    $4\times 4$     & 68.48 & 60.25 & 20.33 & 56.57 & 57.01 \\
    $8\times 8$     & 72.36 & 66.85 & 23.56 & 59.74 & 61.23 \\
    $10\times 10$   & \textbf{74.20} & \textbf{68.33} & \textbf{24.68} & \textbf{62.42} & \textbf{63.60} \\
    $12\times 12$   & 74.14 & 67.56 & 23.90 & 62.02 & 63.50 \\
    \specialrule{.1em}{.05em}{.05em} 
 \end{tabular}}
\label{tab:ablation_q}
\end{table}

\begin{figure}[t]
\vspace{-0.05in}
  \begin{center}
     \includegraphics[width=0.996\linewidth]{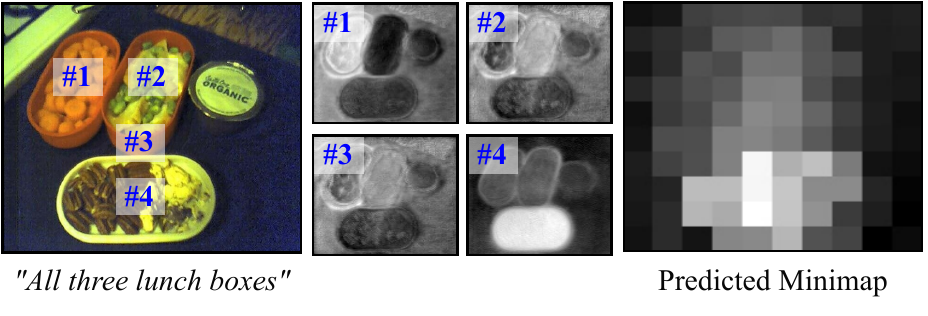}
  \end{center}
 \vspace{-0.25in}
  \caption{Visualization of the predicted minimap \& region masks.}
  \vspace{-0.2in}
  \label{fig:vis_rm}
\end{figure}
\textbf{Number of regions $P$.} 
Smaller $P$ leads to coarser regions, which is not good for capturing fine-grained attributes, while larger $P$ costs more resources and decreases the area of each region, making relationship learning difficult. We do experiments on the selection of $P$ in \cref{tab:ablation_q} to find the optimized $P$. The model's performance improves as $P$ increases until $10$, which is selected as our setting.
In \cref{fig:vis_rm}, we visualize the predicted minimap $x_r$ and region maps $M_r$. $x_r$ displays a rough target probability of each region, showing the effectiveness of minimap supervision. We also see that the region masks capture the spatial correlation of the corresponding regions. With flexible region size and shape, each region mask 
contains not only the instance of this region but also other instances with strong relationships. For example, region \#4 is located inside the bottom lunch box, but as the input expression tells that all three boxes are targets, the top two also cause some responses in the output mask of region \#4.

\begin{table}[t]
  \renewcommand\arraystretch{1.05}
  \centering
  \footnotesize
  \caption{Comparison on gRefCOCO dataset.}
  \vspace{-3.6mm}
  \centering
     \setlength{\tabcolsep}{1.8mm}{\begin{tabular}{l|cl|cl|cl}
      \specialrule{.1em}{.05em}{.05em} 
      \multirow{2}{*}{Methods}& \multicolumn{2}{c|}{val} & \multicolumn{2}{c|}{testA} & \multicolumn{2}{c}{testB} \\
              & cIoU   & gIoU  & cIoU   & gIoU  & cIoU   & gIoU  \\
        \hline\hline
        MattNet~\cite{yu2018mattnet}   & 47.51 & 48.24 & 58.66 & 59.30 & 45.33 & 46.14  \\
        LTS~\cite{jing2021locate}   & 52.30 & 52.70 & 61.87 & 62.64 & 49.96 & 50.42 \\
        VLT~\cite{ding2021vision}   & 52.51 & 52.00 & 62.19 & 63.20 & 50.52 & 50.88 \\
        {CRIS}~\cite{wang2022cris}  & 55.34 & 56.27 & 63.82 & 63.42 & 51.04 & 51.79 \\
        LAVT~\cite{yang2021lavt}  & 57.64 & 58.40 & 65.32 & 65.90 & 55.04 & 55.83 \\
        \hline
        VLT+ReLA  & 58.65 & 59.43 & 66.60 & 65.35 & 56.22 & 57.36 \\
        LAVT+ReLA  & 61.23 & 61.32 & 67.54 & 66.40 & 58.24 & 59.83 \\
        \hline
        \textbf{ReLA} (ours)  & \textbf{62.42} & \textbf{63.60} & \textbf{69.26} & \textbf{70.03} & \textbf{59.88} & \textbf{61.02} \\
        \specialrule{.1em}{.05em}{.05em} 
\end{tabular}}
\vspace{-3mm}
  \label{tab:results_gres}
\end{table}

\begin{table}[t]
    \renewcommand\arraystretch{1.05}
    \centering
    \footnotesize
    \caption{No-target results comparison on gRefCOCO dataset.}
    \vspace{-3.6mm}
    \centering
       \setlength{\tabcolsep}{1.5mm}{\begin{tabular}{l|cc|cc|cc}
        \specialrule{.1em}{.05em}{.05em} 
        \multirow{2}{*}{Methods}& \multicolumn{2}{c|}{val} & \multicolumn{2}{c|}{testA} & \multicolumn{2}{c}{testB} \\
                & \ntacc   & \tacc  & \ntacc   & \tacc  & \ntacc   & \tacc \\
          \hline\hline
          MattNet~\cite{yu2018mattnet}   & 41.15 & 96.13 & 44.04 & 97.56 & 41.32 & 95.32 \\
          VLT~\cite{ding2021vision}   & 47.17 & 95.72 & 48.74 & 95.86 & 47.82 & 94.66 \\
          LAVT~\cite{yang2021lavt}  & 49.32 & 96.18 & 49.25 & 95.08 & 48.46 & 95.34 \\
          \hline
          \textbf{ReLA}-50pix  & 49.96 & 96.28 & 51.36 & 96.35 & 49.24 & 95.02 \\
          \textbf{ReLA}  & \textbf{56.37} & \textbf{96.32} & \textbf{59.02} & \textbf{97.68} & \textbf{58.40} & \textbf{95.44} \\
          \specialrule{.1em}{.05em}{.05em} 
  \end{tabular}}
  \vspace{-0.16in}
\label{tab:results_gres_nt}
\end{table}

\begin{table*}[ht]
\renewcommand\arraystretch{1.0} 
     \centering
     \footnotesize
     \caption{Results on classic RES in terms of cIoU. U: UMD split. G: Google split.}
    \vspace{-3.6mm}
     \setlength{\tabcolsep}{2.7mm}{\begin{tabu}{l|c|c|ccc|ccc|ccc}
      \specialrule{.1em}{.05em}{.05em} 
        \multirow{2}{*}{Methods} &\multirow{2}{*}{\shortstack{Visual\\Encoder}} & \multirow{2}{*}{\shortstack{Textual\\Encoder}}&\multicolumn{3}{c|}{RefCOCO} & \multicolumn{3}{c|}{RefCOCO+} & \multicolumn{3}{c}{G-Ref} \\
          & & & val   & test A & test B & val   & test A & test B & val$_\text{(U)}$   & test$_\text{(U)}$  & val$_\text{(G)}$\\
        \hline
        \hline
        MCN~\cite{luo2020multi}        &Darknet53& bi-GRU &62.44 & 64.20 & 59.71 & 50.62 & 54.99 & 44.69 & 49.22 & 49.40 & -     \\
        {VLT} \cite{ding2021vision} &Darknet53& bi-GRU & 67.52 & 70.47 & 65.24 & 56.30 & 60.98 & 50.08 & 54.96 & 57.73 & 52.02 \\
        {ReSTR}~\cite{kim2022restr} &ViT-B& Transformer &67.22 & 69.30 & 64.45 & 55.78 & 60.44 & 48.27 & -     & -     & 54.48 \\
        {CRIS}~\cite{wang2022cris}& CLIP-R101& CLIP &70.47 & 73.18 & 66.10 & 62.27 & 68.08 & 53.68 & 59.87 & 60.36     & - \\
        {LAVT}~\cite{yang2021lavt}&{Swin-B}& BERT &72.73 & {75.82} & 68.79 & 62.14 & 68.38 & 55.10 & 61.24 & 62.09 & 60.50 \\
        {{VLT~\cite{VLTPAMI}}} &{Swin-B}& BERT&{72.96} & {75.96} & {69.60} & {63.53} & {68.43} & {56.92} & {63.49} & \textbf{66.22} & \textbf{62.80} \\
        \hline
        {\textbf{ReLA} (ours)}&{Swin-B}& BERT & \textbf{73.82} & \textbf{76.48} & \textbf{70.18} & \textbf{66.04} & \textbf{71.02} & \textbf{57.65} & \textbf{65.00} & {65.97} & {62.70} \\
        \specialrule{.1em}{.05em}{.05em} 
     \end{tabu}}%
    \vspace{-0.15in}
     \label{tab:results_res}%
\end{table*}%

\subsection{Results on GRES}

\textbf{Comparison with state-of-the-art RES methods.} In \cref{tab:results_gres}, we report the results of classic RES methods on gRefCOCO.
We re-implement these methods using the same backbone as our model and train them on gRefCOCO. 
For one-stage networks, output masks with less than 50 positive pixels are cleared to all-negative, for better no-target identification.
For the two-stage network MattNet~\cite{yu2018mattnet}, we let the model predict a binary label for each instance that indicates whether this candidate is a target, then merge all target instances. 
As shown in \cref{tab:results_gres}, these classic RES methods do not perform well on gRefCOCO that contains multi-target and no-target samples. Furthermore, to better verify the effectiveness of explicit modeling, we add our ReLA on VLT~\cite{ding2021vision} and LAVT~\cite{yang2021lavt} to replace the decoder part of them. From \cref{tab:results_gres}, our explicit relationship modeling greatly enhances model's performance. \Eg, adding ReLA improves the cIoU performance of the LAVT by more than $4\%$ on the val set.

In \cref{tab:results_gres_nt}, we test the no-target identification performance. 
As shown in the table, \tacc~of all methods are mostly higher than $95\%$, showing that our gRefCOCO does not significantly affect the model's targeting performance while being generalized to no-target samples. 
But from \ntacc of classic RES methods, we see that even being trained with
no-target samples, it is not satisfactory to identify no-target samples solely based on the output mask. We also tested our model with the no-target classifier disabled and only use the positive pixel count in the output mask to identify no-target samples (``{ReLA}-50pix'' in \cref{tab:results_gres_nt}). The performance is similar to other methods. This shows that a dedicated no-target classifier is desired. 
However, although our \ntacc is higher than RES methods, there 
are still around $40\%$ of no-target samples are missed. 
We speculate that this is because many no-target expressions are very deceptive and similar with real instances in the image. We believe that no-target identification will be one of our key focus on the future research for the GRES task.

\begin{figure}[t]
  \vspace{-0.04in}
  \begin{center}
     \includegraphics[width=0.996\linewidth]{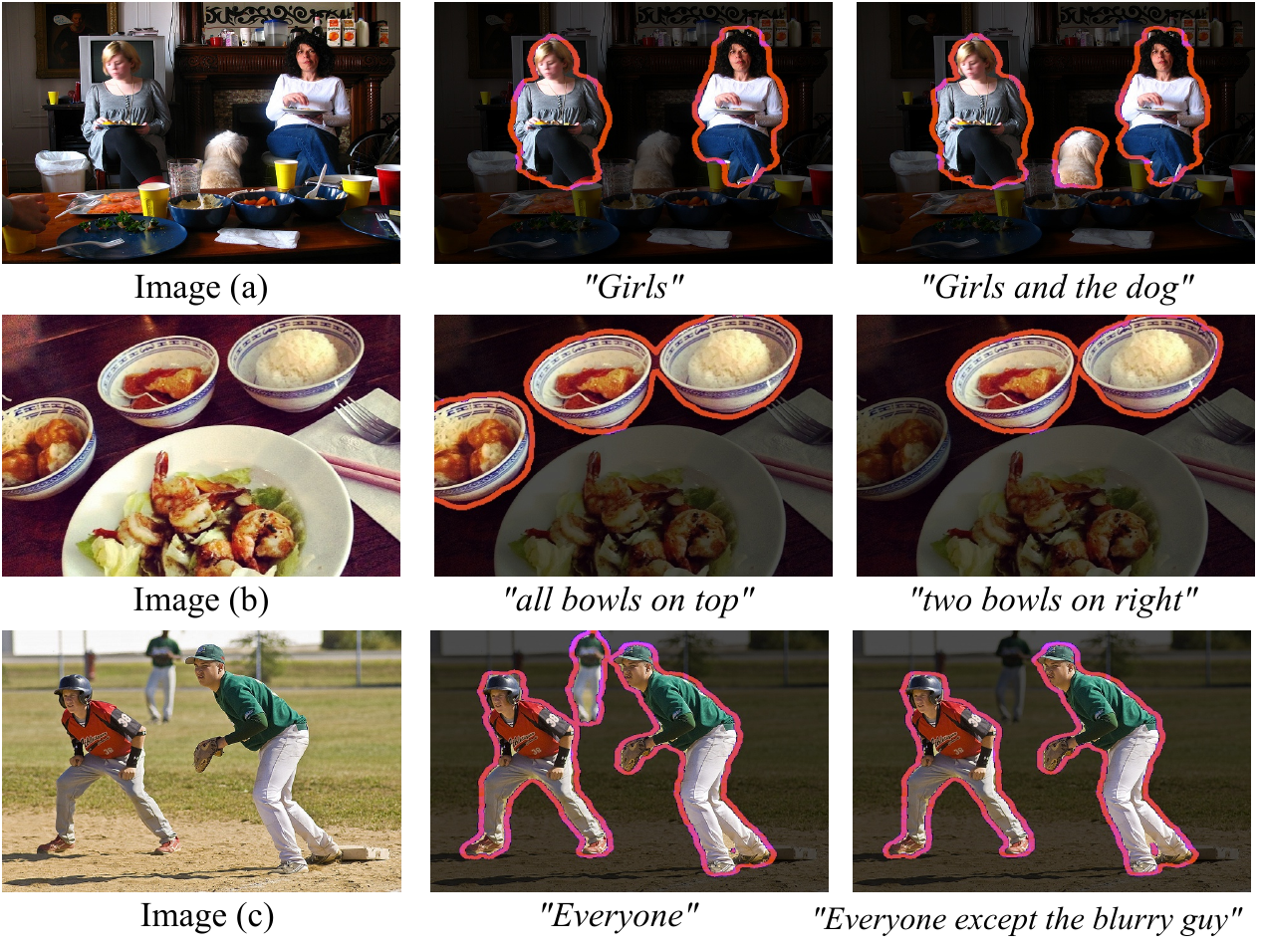}
  \end{center}
 \vspace{-0.24in}
  \caption{Example results of our method on gRefCOCO dataset.}
  \vspace{-0.19in}
  \label{fig:gRefCOCO-demo}
\end{figure}

\textbf{Qualitative results.} Some qualitative examples of our model on the val set of gRefCOCO are shown in \cref{fig:gRefCOCO-demo}. In Image (a), our model can detect and precisely segment multiple targets of the same category (\textit{``girls''}) or different categories (\textit{``girls and the dog''}), showing the strong generalization ability. Image (b) uses counting words (\textit{``two bowls''}) and shared attributes (\textit{``on right''}) to describe a set of targets. Image (c) has a compound sentence showing that our model can understand the excluding relationship: \textit{``except the blurry guy''} and makes a good prediction. 

\begin{figure}[t]
\vspace{-3mm}
  \begin{center}
     \includegraphics[width=0.996\linewidth]{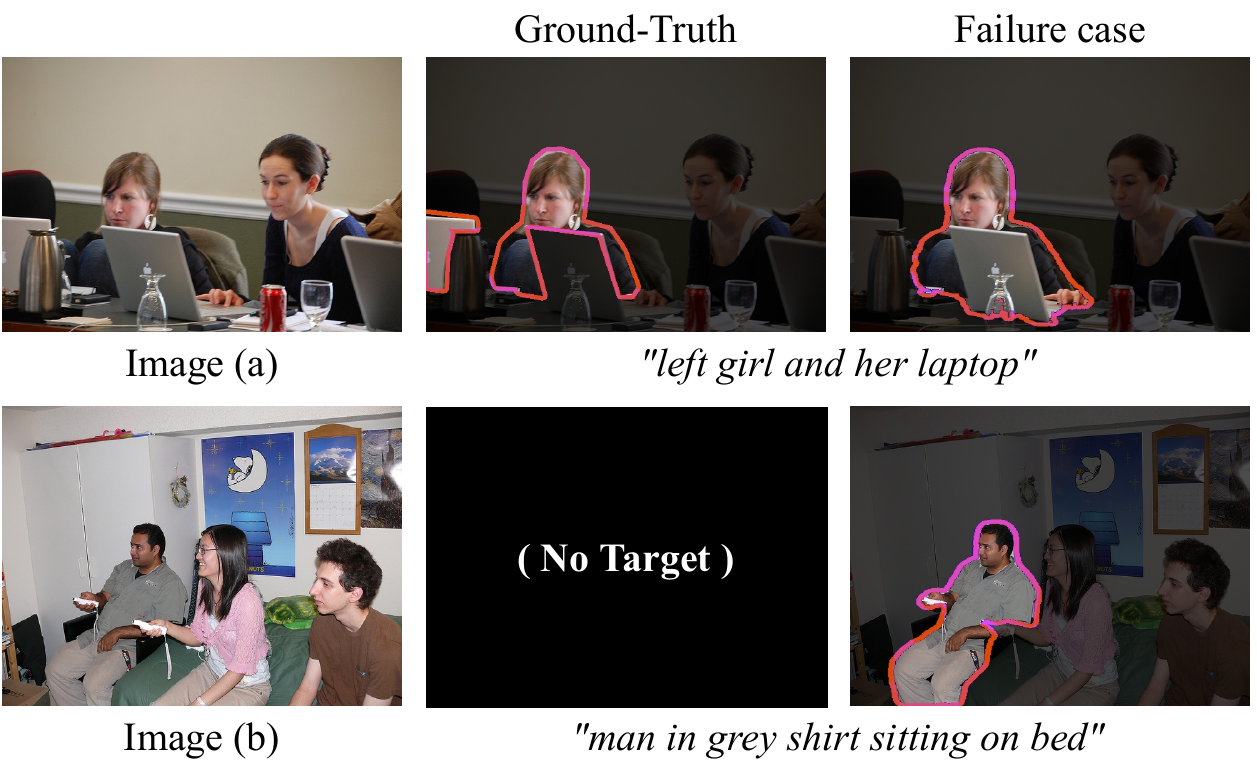}
  \end{center}
 \vspace{-0.2in}
  \caption{Failure cases of our method on gRefCOCO dataset.}
  \vspace{-0.2in}
  \label{fig:gRefCOCO-failure}
\end{figure}

\textbf{Failure cases \& discussion.} We show some failure cases of our method in \cref{fig:gRefCOCO-failure}. Image (a) introduces a possession relationship: \textit{``left girl and \textbf{her} laptop''}. This is a very deceptive case. In the image, the laptop in center is more dominant and closer to the left girl than the left one, so the model highlighted the center laptop as \textit{``her laptop''}. Such a challenging case requires the model to have a profound understanding of all objects, and a contextual comprehension of the image and expression. In the second case, the expression is a no-target expression, referring to \textit{``man in gray shirt sitting on bed''}. In the image, there is indeed a sitting person in grey shirt, but he is sitting on a black chair very close to the bed. This further requires the model to look into the fine-grained details of all objects, and understand those details with image context.

\subsection{Results on Classic RES}

We also evaluate our method on the classic RES task and report the results in \cref{tab:results_res}. In this experiment, our model strictly follows the setting of previous methods~\cite{ding2021vision,yang2021lavt} and is only trained on the RES datasets. As shown in \cref{tab:results_res}, the proposed approach ReLA outperforms other methods on classic RES. Our performance is consistently higher than the state-of-the-art LAVT~\cite{yang2021lavt} with a margin of 1\%$\sim$4\% on three datasets. 
Although the performance gain of our proposed method over other methods on classic RES is lower than that on GRES, the results show that the explicit relationship modeling is also beneficial to classic RES. More results are reported in Supplementary Materials.

\section{Conclusion}

We analyze and address the limitations of the classic RES task, \ie, it cannot handle multi-target and no-target expressions. Based on that, a new benchmark, called Generalized Referring Expression Segmentation (GRES), is defined to allow an arbitrary number of targets in the expressions. To support the research on GRES, we construct a large-scale dataset gRefCOCO. We propose a baseline method ReLA for GRES to explicitly model the relationship between different image regions and words, which consistently achieves new state-of-the-art results on the both classic RES and newly proposed GRES tasks.
The proposed GRES greatly reduces the constraint to the natural language inputs, increases the application scope to the cases of multiple instances and no right objects in image, and opens possible new applications such as image retrieval.

{\small
\bibliographystyle{ieee_fullname}
\bibliography{egbib}
}

\end{document}